# Indian Sign Language Recognition Using Eigen Value Weighted Euclidean Distance Based Classification Technique

Joyeeta Singha
Dept. of Electronics and Communication Engineering
Assam Don Bosco University
Guwahati, India

Karen Das
Dept. of Electronics and Communication Engineering
Assam Don Bosco University
Guwahati, India

*Abstract*—Sign Language Recognition is one of the most growing fields of research today. Many new techniques have been developed recently in these fields. Here in this paper, we have proposed a system using Eigen value weighted Euclidean distance as a classification technique for recognition of various Sign Languages of India. The system comprises of four parts: Skin Filtering, Hand Cropping, Feature Extraction and Classification. 24 signs were considered in this paper, each having 10 samples, thus a total of 240 images was considered for which recognition rate obtained was 97%.

## I. INTRODUCTION

Sign Language is a well-structured code gesture, every gesture has meaning assigned to it. Sign Language is the only means of communication for deaf people. With the advancement of science and technology many techniques have been developed not only to minimize the problem of deaf people but also to implement it in different fields. Many research works related to Sign languages have been done as for example the American Sign Language, the British Sign Language, the Japanese Sign Language, and so on. But very few works has been done in Indian Sign Language recognition till date.

Finding an experienced and qualified interpreters every time is a very difficult task and also unaffordable. Moreover, people who are not deaf, never try to learn the sign language for interacting with the deaf people. This becomes a cause of isolation of the deaf people. But if the computer can be programmed in such a way that it can translate sign language to text format, the difference between the normal people and the deaf community can be minimized.

We have proposed a system which is able to recognize the various alphabets of Indian Sign Language for Human-Computer interaction giving more accurate results at least possible time. It will not only benefit the deaf and dumb people of India but also could be used in various applications in the technology field.

## II. LITERATURE REVIEW

Different approaches have been used by different researchers for recognition of various hand gestures which were implemented in different fields. Some of the approaches were vision based approaches, data glove based approaches, soft computing approaches like Artificial Neural Network, Fuzzy logic, Genetic Algorithm and others like PCA, Canonical Analysis, etc. The whole approaches could be divided into three broad categories- Hand segmentation approaches, Feature extraction approaches and Gesture recognition approaches. Few of the works have been discussed in this paper.

Many researchers [1-11] used skin filtering technique for segmentation of hand. This technique separated the skin colored pixels from the non-skin colored pixels, thus extracting the hand from the background. Fang [12] used Adaptive Boost algorithm which could not only detect single hand but also the overlapped hands. In [13-15] external aid like data gloves, color gloves were used by the researchers for segmentation purpose.

Saengsri [13] in his paper Thai Sign Language Recognition used '5DT Data Glove 14 Ultra' data glove which was attached with 14 sensors- 10 sensors on fingers and rest 4 sensors between the fingers which measures flexures and abductions respectively. But accuracy rate was 94%. Kim [14] used 'KHU-1' data glove which comprises of 3 accelerometer sensor, a Bluetooth and a controller which extracted features like joints of hand. He performed the experiment for only 3 gestures and the process was very slow. Weissmann [15] used Cyberglove which measured features like thumb rotation, angle made between the neighboring fingers and wrist pitch. Limitations were that the system could recognize only single hand gestures.

There have been wide approaches for feature extraction like PCA, Hit-Miss Transform, Principle Curvature Based Region detector (PCBR), 2-D Wavelet Packet Decomposition (WPD) etc. In [1][16-18] Principal Component Analysis (PCA) was used for extracting features for recognition of various hand gestures. Kong [16] segmented the 3-D images into lines and curves and then PCA was used to determine features like direction of motion, shape, position and size.

Lamar [17] in his paper for American and Japanese alphabet recognition used PCA for extracting features like position of the finger, shape of the finger and direction of the image described by the mean, Eigen values and Eigen vectors respectively. The limitations were accuracy rate obtained was 93% which was low and the system could recognize gestures of only single hand. Kapuscinski [2] proposed Hit-Miss transform for extracting features like orientation, hand size by computing the central moments. Accuracy rate obtained was 98% but it lacks proper Skin filtering with changes in illumination. Generic Fourier descriptor and Generic Cosine Descriptor is used [19] for feature extraction as it is rotation invariant, translation invariant and scale invariant. Rotation of the input hand image leads to shifting of hand in polar space. Rotation invariance is obtained by only considering the magnitude of the Fourier coefficient. While using centroid as the origin translational invariance is achieved and finally ratio of magnitude to area scale invariance is obtained. Only 15 different hand gestures were considered in this paper. Rekha [9] extracted the texture, shape, finger features of hand in the form of edges and lines by PCBR detector which otherwise is a very difficult task because of change in illumination, color and scale. Accuracy rate obtained was 91.3%.

After the features were extracted, proper classifier were used to recognize the gestures. There are various gesture recognition approaches used by different researchers like Support Vector Machines, Artificial Neural Network (ANN), Genetic Algorithm (GA), Fuzzy Logic, Euclidean distance, Hidden Markov Model (HMM), etc. [13][17] used ANN for recognizing gestures. Saengsri [13] used Elman Back Propagation Neural Network (ENN) algorithm which consisted of input layer with 14 nodes similar to the sensors in the data glove, output layer with 16 nodes equal to the number of symbols and hidden layer with 30 nodes which is just the total of input and output nodes. Gesture was recognized by identifying the maximum value class from ENN. Recognition rate obtained was 94.44%. Difficulty faced in this paper was it considered only single gestured signs. Lamar [17] used ANN which comprises of input layer with 20 neurons, hidden and output layer each with 42 neurons. Back propagation algorithm was used and after the training of the neural network one output neuron was achieved, thus giving the proper recognized gesture. Gopalan [1] used Support Vector Machine for classification purpose. The linearly non separable data becomes separable when SVM was used as the data was projected to higher dimensional space, thus reducing error. Kim [20] in his paper of Recognition of Korean Sign Language used Fuzzy logic. Fuzzy sets were considered where each set were the various speeds of the moving hand. They were mathematically given by ranges like small, medium, negative medium, large, positive large, etc. Accuracy rate obtained was 94% and difficulty faced by them was heavy computation.

We have thus proposed a system that could overcome the difficulties faced by various. Our proposed system was able to recognize two hand gestures with an improved accuracy rate of 97%. Moreover, experiment was carried out with bare hands and computational time was also less thus removing the difficulties faced by use of the hand gloves with sensors.

III. THEORETICAL BACKGROUND

A. *Eigen value and Eigen vector*

Eigen values and Eigen vectors are a part of linear transformations. Eigen vectors are the directions along which the linear transformation acts by stretching, compressing or flipping and Eigen values gives the factor by which the compression or stretching occurs. In case of analysis of data, the Eigen vectors of the covariance are being found out. Eigenvectors are set of basis function which describes variability of data. And Eigen vectors are also a kind of coordinate system for which the covariance matrix becomes diagonal for which the new coordinate system is uncorrelated. The more the Eigen vectors the better the information obtained from the linear transformation. Eigen values measures the variance of data of new coordinate system. For compression of the data only few significant Eigen values are being selected which reduces the dimension of the data allowing the data to get compressed. Mathematically, it is explained in (1).

If $x$ is a one column vector with $n$ rows and $A$ is a square matrix with $n$ rows and columns, then the matrix product $Ax$ will result in vector $y$. When these two vectors are parallel, $Ax = \lambda x$, ($\lambda$ being any real number) then $x$ is an eigenvector of $A$ and the scaling factor $\lambda$ is the respective eigenvalue.

$$\begin{bmatrix} x_1 \\ x_2 \\ \vdots \\ x_n \end{bmatrix} \rightarrow \begin{bmatrix} A_{11} & A_{12} & \cdots & A_{1n} \\ A_{21} & A_{22} & \ddots & A_{2n} \\ \vdots & \vdots & & \vdots \\ A_{n1} & A_{n2} & \cdots & A_{nn} \end{bmatrix} \begin{bmatrix} x_1 \\ x_2 \\ \vdots \\ x_n \end{bmatrix} = \begin{bmatrix} y_1 \\ y_2 \\ \vdots \\ y_n \end{bmatrix} \quad (1)$$

IV. PROPOSED SYSTEM

The block diagram of the proposed system is given in Fig. 1 which comprises of mainly four phases: Skin filtering, Hand cropping, Feature Extraction and Classification.

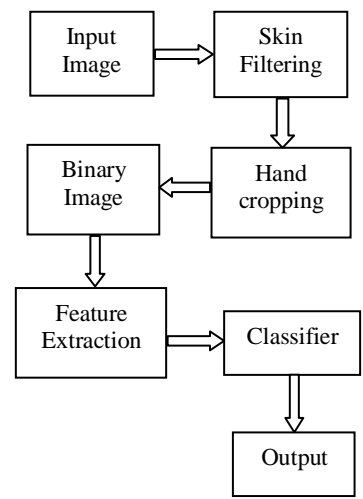

Fig. 1. Block Diagram Of The Proposed System

In our proposed system, we have considered 24 alphabets of Indian sign language, each with 10 samples thus a total of 240 images captured by camera. Some of the database images have been shown for each alphabet in Fig. 2.

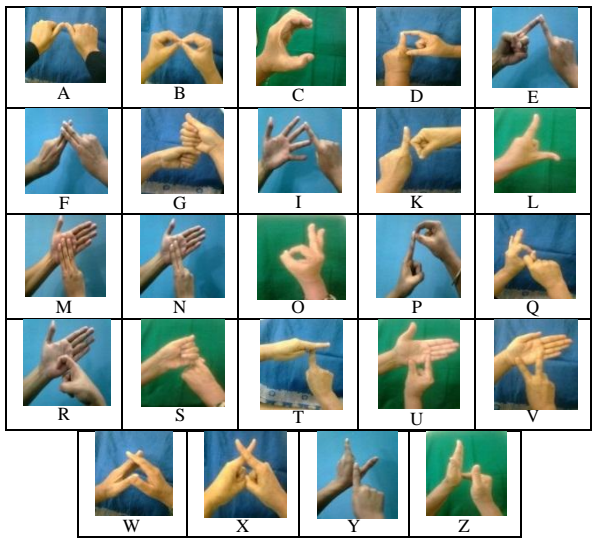

Fig. 2. Some of the database images considered for the proposed system

### A. Skin Filtering

The first phase for our proposed system is the skin filtering of the input image which extracts out the skin colored pixels from the non-skin colored pixels. This method is very much useful for detection of hand, face etc. The steps carried out for performing skin filtering is given in Fig. 3.

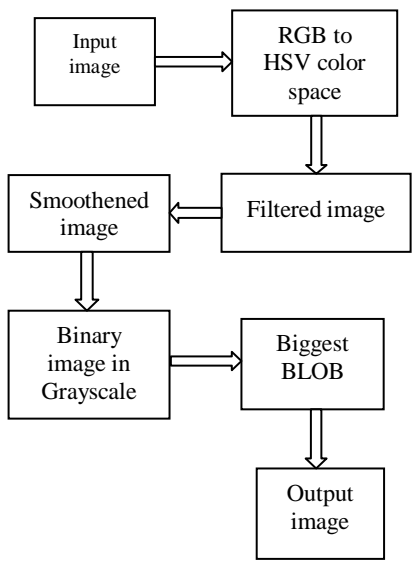

Fig. 3. Basic Block diagram of Skin Filtering

The input RGB image is first converted to the HSV image. The motive of performing this step is RGB image is very sensitive to change in illumination condition. The HSV color space separates three components: Hue which means the set of pure colors within a color space, Saturation describing the grade of purity of a color image and Value giving relative lightness or darkness of a color. The following Fig. 4 shows the different components of HSV color model.

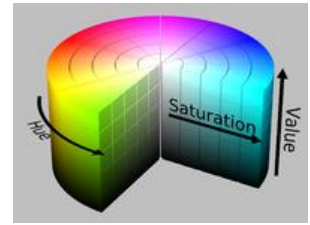

Fig. 4. HSV color model

In our proposed system, the RGB is converted to HSV color model by the following mathematical calculations:

$$H = \begin{cases} 60\left(\frac{G-B}{\delta}\right) & if\ MAX = R \\ 60\left(\frac{B-R}{\delta} + 2\right) & if\ MAX = G \\ 60\left(\frac{R-G}{\delta} + 4\right) & if\ MAX = B \\ not\ defined & if\ MAX = 0 \end{cases} \quad (2)$$

$$s = \begin{cases} \frac{\delta}{MAX} & if\ MAX \neq 0 \\ 0 & if\ MAX = 0 \end{cases} \quad (3)$$

where $\delta = (MAX - MIN)$, $MAX = \max(R, G, B)$ and $MIN = \min(R, G, B)$.

Then the HSV image is filtered and smoothened and finally we get an image which comprises of only skin colored pixels. Now, along with the hand other objects in the surroundings may also have skin-color like shadows, wood, dress etc. Thus to eliminate these, we take the biggest binary linked object (BLOB) which considers only the region comprising of biggest linked skin-colored pixels. Results obtained after performing skin filtering is given in Fig. 5.

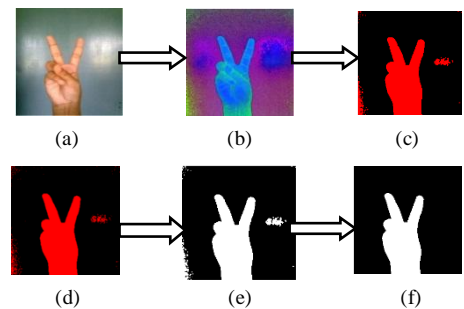

Fig. 5. a) RGB image, b) HSV image, c) Filtered image, d) Smoothened image, e) Binary image in grayscale, f) Biggest BLOB.

### B. Hand Cropping

Next phase is the cropping of hand. For recognition of different gestures, only hand portion till wrist is required, thus the unnecessary part is clipped off using this hand cropping technique. Significance of using this hand cropping is we can detect the wrist and hence eliminate the undesired region. And once the wrist is found the fingers can easily be located as it will lie in the opposite region of wrist. The steps involved in this technique are as follows.

- The skin filtered image is scanned from all direction left, right, top, bottom to detect the wrist of the hand. Once the wrist is detected its position can be easily found out.
- Then the minimum and maximum positions of the white pixels in the image are found out in all other directions. Thus we obtain $X_{min}$, $Y_{min}$, $X_{max}$, $Y_{max}$, one of which is the wrist position.
- Then the image is cropped along these coordinates as used in [5].

Few images have been shown in Fig. 6 after performing hand cropping.

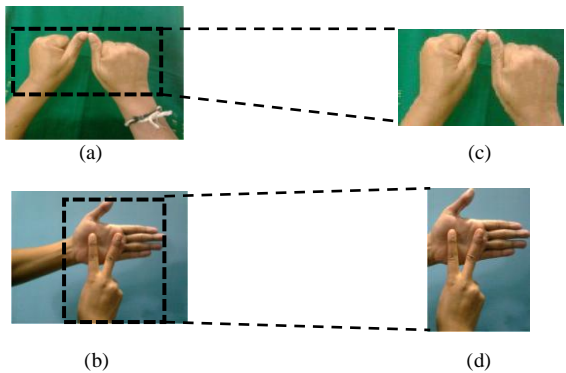

Fig. 6. Hand cropping: (a) and (b) showing the input image, (c) and (d) showing cropped image respectively.

*C. Feature Extraction*

After the desired portion of the image is being cropped, feature extraction phase is carried out. Here, Eigen values and Eigen vectors are found out from the cropped image. The mathematical steps for finding out Eigen values and Eigen vectors in our proposed system are:

- The input data is assumed to be X. Here, in our paper cropped image has been taken as the input image having a dimension 50 by 50.
- The mean of the above vector X is found out as
$$M = E\{X\} \quad (4)$$
- Then the covariance matrix C of the above input vector X was found out. Mathematically, it was given by
$$C = E\{(X - M)(X - M)'\} \quad (5)$$
- The Eigen vectors and the Eigen values are computed from the covariance matrix C.
- Finally the Eigen vectors are arranged in such a way that the corresponding Eigen values is in the decreasing order.

In our project, only five significant Eigen vectors out of 50 has been considered because the Eigen values were very small after this and so can be neglected. This provides advantages like data compression, data dimension reduction without much loss of information, reducing the original variables into a lower number of orthogonal or non-correlated synthesized variables.

*D. Classifier*

Classifier was needed in order to recognize various hand gestures. In our paper, we have designed a new classification technique that is Eigen value weighted Euclidean distance between Eigen vectors which involved two levels of classification.

- Classification based on Euclidean Distance: Euclidean distance was found out between the Eigen vectors of the test image and the corresponding Eigen vectors of the database image. As five Eigen vectors were considered, we get five Euclidean distances for each database image and then the minimum of each was found out. Mathematically,

$$E.D. = \sqrt{\sum_{n=1}^{m}(EV1(n) - EV2(n))^2} \quad (6)$$

where EV1 represents the Eigen vectors of the test image and EV2 represents the Eigen vectors of the database image.

- Classification based on Eigen value weighted Euclidean distance: The difference of Eigen values of the test image and the Eigen values of the database image was found out. Then, it was multiplied with the Euclidean Distance obtained in the first level of classification given as C2 in equation below. Then sum of results obtained for each image were added and minimum of them was considered to be the recognized symbol. Mathematically,

$$C2 = (E.D.) * |E1 - E2| \quad (7)$$

where E1 and E2 are the Eigen values of the test images and database images respectively.

V. RESULTS AND DISCUSSIONS

Different images were tested and found that the new technique of classification was found to show 97% accuracy. Some images tested with other database images are given in the following table where 2 levels of classification were used to identify the gestures. Table I shows the Level 1 classification experimented for different test images and Table II shows the level 2 classification.

A comparison between the first level and second level of classification is being made in Table III and it is seen that the success rate has improved from 87% to 97% with the use of the Eigen value weighted Euclidean distance between Eigen vectors as a classification technique.

TABLE I. CLASSIFICATION BASED ON EUCLIDEAN DISTANCE

| Test image | Image in database | Euclidean distance with 1st Eigen vector | Euclidean distance with 2nd Eigen vector | Euclidean distance with 3rd Eigen vector | Euclidean distance with 4th Eigen vector | Euclidean distance with 5th Eigen vector | Recognized symbol |
|---|---|---|---|---|---|---|---|
| 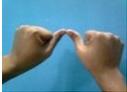 | A | **0.1249** | 1.5691 | 1.2558 | **0.4792** | **0.8158** | "A" |
| | B | 0.5533 | 0.5956 | 1.7043 | 1.4447 | 1.4507 | |
| | C | 0.7618 | 0.7394 | 1.0156 | 1.3916 | 1.3854 | |
| | D | 0.9854 | 1.2047 | 0.9849 | 1.5242 | 1.6026 | |
| | E | 0.5963 | 0.7418 | 1.6339 | 1.5727 | 1.6066 | |
| | F | 0.9521 | 1.0793 | 1.4544 | 1.5081 | 1.2504 | |
| | G | 1.1609 | 1.6549 | 1.7979 | 1.3987 | 1.7241 | |
| | I | 0.8485 | 0.8528 | 0.8169 | 1.2077 | 1.3014 | |
| | K | 0.9268 | 0.9928 | 0.5444 | 1.2782 | 1.0812 | |
| | L | 0.6364 | 1.9378 | 0.6811 | 0.8108 | 1.6678 | |
| | M | 0.3860 | 1.6395 | 1.4842 | 1.7437 | 1.3255 | |
| | N | 0.4770 | 1.1493 | 1.4225 | 1.7111 | 1.4469 | |
| | O | 0.6612 | 0.8577 | 1.4895 | 1.5931 | 1.3063 | |
| | P | 1.0740 | 1.0917 | 1.0965 | 1.3409 | 1.1151 | |
| | Q | 1.3458 | 1.4588 | 0.8631 | 1.7031 | 1.3686 | |
| | R | 1.1635 | 1.2585 | 1.1592 | 1.0778 | 1.6516 | |
| | S | 1.5031 | 1.1822 | 1.7871 | 0.8983 | 1.6370 | |
| | T | 0.9091 | 1.0428 | 0.8999 | 1.1844 | 1.2316 | |
| | U | 0.4152 | 1.0505 | 1.0741 | 1.1402 | 1.2867 | |
| | V | 0.4867 | 1.1147 | 1.3363 | 1.0399 | 1.4031 | |
| | W | 1.5046 | 1.2852 | 1.2904 | 1.6789 | 1.3340 | |
| | X | 1.4303 | 1.4346 | 1.6386 | 1.6693 | 1.3324 | |
| | Y | 1.5174 | 1.4646 | 1.1740 | 1.4543 | 1.5043 | |
| | Z | 1.4874 | 1.3243 | 0.9958 | 1.3852 | 1.4072 | |

TABLE II. CLASSIFICATION BASED ON EIGEN VALUE WEIGHTED EUCLIDEAN DISTANCE

| Test image | Image in database | Eigen value weighted E.D. (1st Eigen values) | Eigen value weighted E.D. (2nd Eigen values) | Eigen value weighted E.D. (3rd Eigen values) | Eigen value weighted E.D. (4th Eigen values) | Eigen value weighted E.D. (5th Eigen values) | Sum | Recognized symbol |
|---|---|---|---|---|---|---|---|---|
| 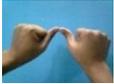 | A | 0.0536 | 2.6397 | 0.1272 | 0.2379 | 0.1277 | **3.1861** | "A" |
| | B | 2.0381 | 1.2870 | 0.4815 | 0.8599 | 0.2717 | 4.9382 | |
| | C | 0.9585 | 1.4476 | 0.0278 | 1.5120 | 0.0891 | 4.0350 | |
| | D | 2.4494 | 0.6792 | 0.0666 | 1.3786 | 0.0126 | 4.5864 | |
| | E | 0.7930 | 1.7588 | 1.3505 | 2.1545 | 0.0951 | 6.1519 | |
| | F | 1.5441 | 1.3513 | 1.2021 | 1.5735 | 0.0472 | 5.7182 | |
| | G | 0.6581 | 4.8827 | 3.2120 | 1.9195 | 0.3106 | 10.9829 | |
| | I | 3.5201 | 0.3021 | 0.2475 | 0.2014 | 1.0829 | 5.3540 | |
| | K | 7.0507 | 0.2316 | 0.3913 | 1.4674 | 0.0889 | 9.2299 | |
| | L | 2.6877 | 1.3673 | 0.1965 | 0.5769 | 0.5458 | 5.3742 | |
| | M | 0.4923 | 5.9935 | 0.3323 | 2.3351 | 0.0834 | 9.2366 | |
| | N | 0.6266 | 3.4768 | 0.2533 | 2.2766 | 0.0606 | 6.6939 | |
| | O | 1.2939 | 1.9144 | 1.2720 | 1.4887 | 0.3475 | 6.3165 | |
| | P | 0.8887 | 1.4092 | 1.8448 | 1.0864 | 0.3796 | 5.6087 | |
| | Q | 2.4164 | 0.1453 | 0.2031 | 0.5231 | 0.0593 | 3.3472 | |
| | R | 2.3273 | 2.4424 | 0.1145 | 0.8104 | 0.1669 | 5.8615 | |
| | S | 0.2087 | 0.9724 | 2.6583 | 0.3222 | 0.1852 | 4.3468 | |
| | T | 3.0581 | 0.3526 | 1.3615 | 1.7494 | 0.1140 | 6.6356 | |
| | U | 0.8202 | 2.0471 | 0.0857 | 0.3893 | 0.4308 | 3.7731 | |
| | V | 0.2843 | 2.6518 | 1.1536 | 0.2339 | 0.2783 | 4.6019 | |
| | W | 4.0941 | 0.8412 | 0.6467 | 2.4152 | 0.1949 | 8.1921 | |
| | X | 4.5624 | 0.4257 | 0.8330 | 2.8831 | 0.1120 | 8.8162 | |
| | Y | 1.8397 | 0.0949 | 0.5030 | 0.9159 | 0.6398 | 3.9933 | |
| | Z | 3.1385 | 2.5160 | 0.2507 | 1.4153 | 0.0757 | 7.3962 | |

TABLE III. SUCCESS RATES OF TWO LEVELS OF CLASSIFICATION

| Symbol | Number of images experimented | Success rate of Euclidean distance classification | Success rate of Eigen value weighted Euclidean distance classification |
|---|---|---|---|
| A | 10 | 100% | 100% |
| B | 10 | 90% | 100% |
| C | 10 | 100% | 100% |
| D | 10 | 70% | 90% |
| E | 10 | 90% | 100% |
| F | 10 | 90% | 90% |
| G | 10 | 100% | 100% |
| I | 10 | 90% | 100% |
| K | 10 | 90% | 100% |
| L | 10 | 90% | 100% |
| M | 10 | 70% | 80% |
| N | 10 | 70% | 90% |
| O | 10 | 100% | 100% |
| P | 10 | 80% | 90% |
| Q | 10 | 90% | 100% |
| R | 10 | 90% | 100% |
| S | 10 | 100% | 100% |
| T | 10 | 100% | 100% |
| U | 10 | 70% | 90% |
| V | 10 | 70% | 90% |
| W | 10 | 80% | 100% |
| X | 10 | 80% | 100% |
| Y | 10 | 80% | 100% |
| Z | 10 | 100% | 100% |

From the above experiments, we can say that we have designed a system that was able to recognize different alphabets of Indian Sign Language and we have removed difficulties faced by the previous works with improved recognition rate of 97%. The time taken to process an image was 0.0384 seconds. Table IV describes a brief comparative study between our works with the other related works.

## VI. CONCLUSION AND FUTURE WORK

The proposed system was implemented with MATLAB version 7.6 (R2008a) and supporting hardware was Intel® Pentium® CPU B950 @ 2.10GHz processor machine, Windows 7 Home basic (64 bit), 4GB RAM and an external 2 MP camera. A system was designed for Indian Sign Language Recognition. It was able to handle different static alphabets of Indian Sign Languages by using Eigen value weighted Euclidean distance between Eigen vectors as a classification technique. We have tried to improve the recognition rate compared to the previous works and achieved a success rate of 97%. Moreover, we have considered both hands in our paper. As we have performed the experiments with only the static images so out of the 26 alphabets 'H' and 'J' were not considered as they were dynamic gestures. We hope to deal with dynamic gestures in future. Moreover only 240 images were considered in this paper so in future we hope to extend it further.

TABLE IV. COMPARITIVE STUDY BETWEEN OUR WORK AND OTHER APPROACHES

| Name of the technique used | Success Rate | Difficulties faced |
|---|---|---|
| Hit-Miss Operation, HMM [2] | 97.83% | Weak skin color detection |
| PCA, Gabor Filter and SVM [3] | 95.2% | Single hand gesture recognition |
| Perceptual color space [6] | 100% | Dealt with only 5 hand gestures |
| Contour based [8] | 91% | • Use of hand gloves<br>• Single hand gesture recognition |
| ANN based [13] | 94% | Use of data gloves with 13 sensors |
| Kinematic Chain Theory based [14] | 100% | • 3 simple hand gesture recognition<br>• Use of Data gloves<br>• Reduction in computation time |

| Our Work | Success rate | Advantages |
|---|---|---|
| Eigen value weighted Euclidean Distance based | 97% | • Less computation time<br>• Can recognize 2 hand gestures<br>• Performed with bare hands, thus removing difficulties of using gloves<br>• Can recognize 24 different gestures with high success. |

REFERENCES

[1] R. Gopalan and B. Dariush, "Towards a Vision Based Hand Gesture Interface for Robotic Grasping", The IEEE/RSJ International Conference on Intelligent Robots and Systems, October 11-15, 2009, St. Louis, USA, pp. 1452-1459.

[2] T. Kapuscinski and M. Wysocki, "Hand Gesture Recognition for Man-Machine interaction", Second Workshop on Robot Motion and Control, October 18-20, 2001, pp. 91-96.

[3] D. Y. Huang, W. C. Hu, and S. H. Chang, "Vision-based Hand Gesture Recognition Using PCA+Gabor Filters and SVM", IEEE Fifth International Conference on Intelligent Information Hiding and Multimedia Signal Processing, 2009, pp. 1-4.

[4] C. Yu, X. Wang, H. Huang, J. Shen, and K. Wu, "Vision-Based Hand Gesture Recognition Using Combinational Features", IEEE Sixth

International Conference on Intelligent Information Hiding and Multimedia Signal Processing, 2010, pp. 543-546.

[5] J. L. Raheja, K. Das, and A. Chaudhury, "An Efficient Real Time Method of Fingertip Detection", International Conference on Trends in Industrial Measurements and automation (TIMA), 2011, pp. 447-450.

[6] Manigandan M. and I. M Jackin, "Wireless Vision based Mobile Robot control using Hand Gesture Recognition through Perceptual Color Space", IEEE International Conference on Advances in Computer Engineering, 2010, pp. 95-99.

[7] A. S. Ghotkar, R. Khatal, S. Khupase, S. Asati, and M. Hadap, "Hand Gesture Recognition for Indian Sign Language", IEEE International Conference on Computer Communication and Informatics (ICCCI), Jan. 10-12, 2012, Coimbatore, India.

[8] I. G. Incertis, J. G. G. Bermejo, and E.Z. Casanova, "Hand Gesture Recognition for Deaf People Interfacing", The 18th International Conference on Pattern Recognition (ICPR), 2006.

[9] J. Rekha, J. Bhattacharya, and S. Majumder, "Shape, Texture and Local Movement Hand Gesture Features for Indian Sign Language Recognition", IEEE, 2011, pp. 30-35.

[10] L. K. Lee, S. Y. An, and S. Y. Oh, "Robust Fingertip Extraction with Improved Skin Color Segmentation for Finger Gesture Recognition in Human-Robot Interaction", WCCI 2012 IEEE World Congress on Computational Intelligence, June, 10-15, 2012, Brisbane, Australia.

[11] S. K. Yewale and P. K. Bharne, "Hand Gesture Recognition Using Different Algorithms Based on Artificial Neural Network", IEEE, 2011, pp. 287-292.

[12] Y. Fang, K. Wang, J. Cheng, and H. Lu, "A Real-Time Hand Gesture Recognition Method", IEEE ICME, 2007, pp. 995-998.

[13] S. Saengsri, V. Niennattrakul, and C.A. Ratanamahatana, "TFRS: Thai Finger-Spelling Sign Language Recognition System", IEEE, 2012, pp. 457-462.

[14] J. H. Kim, N. D. Thang, and T. S. Kim, "3-D Hand Motion Tracking and Gesture Recognition Using a Data Glove", IEEE International Symposium on Industrial Electronics (ISIE), July 5-8, 2009, Seoul Olympic Parktel, Seoul , Korea, pp. 1013-1018.

[15] J. Weissmann and R. Salomon, "Gesture Recognition for Virtual Reality Applications Using Data Gloves and Neural Networks", IEEE, 1999, pp. 2043-2046.

[16] W. W. Kong and S. Ranganath, "Sign Language Phoneme Transcription with PCA-based Representation", The 9th International Conference on Information and Communications Security(ICICS), 2007, China.

[17] M. V. Lamar, S. Bhuiyan, and A. Iwata, "Hand Alphabet Recognition Using Morphological PCA and Neural Networks", IEEE, 1999, pp. 2839-2844.

[18] O. B. Henia and S. Bouakaz, "3D Hand Model Animation with a New Data-Driven Method", Workshop on Digital Media and Digital Content Management (IEEE Computer Society), 2011, pp. 72-76.

[19] M. Pahlevanzadeh, M. Vafadoost, and M. Shahnazi, "Sign Language Recognition", IEEE, 2007.

[20] J. B. Kim, K. H. Park**,** W. C. Bang, and Z. Z. Bien, "Continuous Gesture Recognition System for Korean Sign Language based on Fuzzy Logic and Hidden Markov Model", IEEE, 2002, pp. 1574-1579.